\documentclass[10pt,twocolumn]{article}
\usepackage[letterpaper,margin=0.75in]{geometry}
\usepackage{amsmath,amssymb,amsfonts}
\usepackage{booktabs}
\usepackage{graphicx}
\usepackage{xcolor}
\usepackage{microtype}
\usepackage[font=small,labelfont=bf]{caption}
\usepackage{float}
\usepackage[numbers,square,sort&compress]{natbib}
\usepackage{titlesec}
\titleformat{\section}{\normalfont\large\bfseries}{\thesection}{0.6em}{}
\titleformat{\subsection}{\normalfont\normalsize\bfseries}{\thesubsection}{0.6em}{}
\titlespacing*{\section}{0pt}{1.4ex plus .2ex}{0.8ex}
\titlespacing*{\subsection}{0pt}{1.0ex plus .2ex}{0.5ex}
\begin{document}

\twocolumn[%
\begin{center}
{\LARGE\bfseries Silent Failures in Physics-Informed Neural Networks:\\[2pt]
Parameter Poisoning and the Limits of Loss-Based Validation\par}
\vspace{1.0em}
{\large David McShannon \qquad Nicholas Dietrich\par}
\vspace{1.3em}
\begin{minipage}{0.93\textwidth}
\rule{\linewidth}{0.5pt}\par\vspace{0.5em}
{\small\noindent\textbf{Abstract.} Physics-informed neural networks (PINNs) embed governing equations in their loss function, enabling mesh-free solutions to partial differential equations. Low training loss is treated as evidence that the learned solution is physically correct. This paper shows that assumption breaks down when encoded physics are incorrect. By perturbing PDE parameters before training, a setting we describe as physics parameter poisoning or parameter misspecification, we produce models that train to low loss but give incorrect answers; we treat the perturbation schedule as sensitivity analysis rather than only as a security threat, and none of our claims requires an adversary. Achieving low residual loss does not discriminate accurate from inaccurate solutions: poisoned models reach losses at or below the clean baseline yet differ by large margins, so driving the residual down is not evidence of physical accuracy. Across three PDE systems (Burgers equation, Navier-Stokes cavity, and convection-diffusion), poisoned models match or beat the clean-model training loss while their solutions differ by up to 71\% in the fixed sweep and up to 128\% under adversarial search; at Cavity Re=400 the poisoned loss falls below the clean baseline. We define a detection difficulty ratio R (solution error divided by training loss) to summarize how invisible the corruption is, though cross-PDE comparison is complicated by differences in loss scale. We test six candidate defenses, none of which reliably detects corruption across all regimes. We propose a post-hoc defense: sweeping the PDE residual loss across parameter values without retraining. The loss minimum recovers the true training parameter without external data, and generalizes across all three PDE systems. The effect holds across five network architectures (8.7K to 133K parameters), is bidirectional, and is confirmed across multiple random seeds.\par}
\vspace{0.5em}\rule{\linewidth}{0.5pt}
\end{minipage}
\end{center}
\vspace{1.4em}
]

\section{Introduction}

Physics-informed neural networks (PINNs) solve partial differential equations by embedding the governing PDE directly into the training loss \citep{raissi2019physics}. PINNs do not require labeled data or mesh generation, and have been applied widely across fluid dynamics, heat transfer, solid mechanics, and inverse problems.

A critical assumption in PINN workflows is that successful training (measured by low PDE residual loss) implies that the learned solution is physically correct. Recent high-profile work has extended this trust to mathematical discovery. \citet{wang2025discovery} used PINNs to characterize unstable singularities in fluid equations related to the Navier-Stokes regularity question, with residual minimization as one component of the validation pipeline.

This paper asks what happens when the physics are incorrect. If the PDE parameters encoded in the loss function are incorrect, whether through misconfiguration, pipeline error, or intentional tampering, the trained model can still reach low loss while producing a solution that is physically incorrect. We call this \emph{physics parameter poisoning}, or equivalently parameter misspecification: the trained model exhibits a silent failure in which training diagnostics look normal but the solution is incorrect. We adopt the poisoning terminology to connect with the machine learning security literature, but the same setup equally describes ordinary misconfiguration or a parameter sensitivity analysis, and none of our claims requires an adversary. Unlike prior work on adversarial input perturbations, which perturb evaluation coordinates at inference time, physics parameter poisoning operates at training time by corrupting the equations themselves, making it invisible to any loss-based monitoring that uses the same corrupted equations.

Our contributions:
\begin{enumerate}
    \item We formalize the physics parameter poisoning threat model for PINNs, distinguishing it from input-space attacks.
    \item We demonstrate silent failures across three PDE systems at two complexity levels, showing that vulnerability is PDE-dependent but architecture-independent.
    \item We introduce the \emph{detection difficulty ratio} as a quantitative metric for poisoning severity and show it varies dramatically across PDE types.
    \item We evaluate six candidate defenses, demonstrating that no single defense suffices across all regimes.
    \item We propose a post-hoc loss landscape sweep that detects poisoning by revealing the true training parameter without retraining or external data, validated across all three PDE systems and multiple seeds.
\end{enumerate}

\section{Related Work}

\emph{PINN robustness.} Several works have examined the reliability of PINNs under perturbation, including sensitivity to errors in boundary and initial condition training data \citep{bajaj2023recipes} and extensions of classical adversarial ML \citep{goodfellow2014explaining, madry2017towards} to the scientific computing setting. Our work characterizes a different failure mode: corruption of the physics itself.

\emph{Data poisoning in ML.} Training-time attacks have been extensively studied in supervised learning \citep{biggio2012poisoning, shafahi2018poison}. In the PINN setting, the analogous attack surface is the physics itself: the PDE coefficients, boundary conditions, and domain geometry that define the loss function.

\emph{PINN failure modes.} The broader PINN literature has identified several failure modes including spectral bias \citep{rahaman2019spectral}, causality violations in time-dependent problems \citep{wang2022causality}, and sensitivity to hyperparameters. Parameter corruption has not been systematically characterized in this setting.

\emph{PINNs for scientific discovery.} \citet{wang2025discovery} used PINNs to characterize fluid singularities, validating through residual evaluation alongside self-similarity and eigenvalue checks; pipelines of this kind still rest in part on the assumption that low residuals indicate correctness.

\emph{Inverse PINNs and parameter recovery.} Inverse PINNs treat PDE parameters as learnable variables and recover them jointly with the solution \citep{raissi2019physics}. Our post-hoc loss sweep is related in spirit: both ask which parameter value best explains the model's behavior. The key difference is that inverse PINNs optimize over the parameter using gradient descent on a trainable network, which in our experiments fails because the network absorbs the data-fitting signal (Section~5.3). Our sweep instead evaluates the residual of a \emph{frozen} network over a discrete parameter grid, which is more robust to the over-parameterization that causes gradient-based recovery to fail.

\section{Threat Model}

We consider a scenario in which a PINN is trained to solve a boundary value problem:
\begin{equation}
    \mathcal{N}[u; \lambda] = 0 \text{ on } \Omega, \quad \mathcal{B}[u] = g \text{ on } \partial\Omega, \quad u(\cdot, 0) = u_0,
\end{equation}
where $u$ is the solution, $\mathcal{N}$ is the differential operator, $\mathcal{B}$ is the boundary operator, and $\lambda$ denotes the physical parameters (e.g., Reynolds number, viscosity, diffusivity). The PINN loss is
\begin{equation}
\begin{split}
    \mathcal{L} = {} & \frac{1}{N_r}\sum_{i=1}^{N_r}|\mathcal{N}[\hat{u}; \lambda](x_i)|^2 + \frac{1}{N_b}\sum_{j=1}^{N_b}|\mathcal{B}[\hat{u}](x_j) - g_j|^2 \\
    & + \frac{1}{N_0}\sum_{k=1}^{N_0}|\hat{u}(x_k, 0) - u_{0,k}|^2,
\end{split}
\end{equation}
where $\hat{u}$ is the neural network approximation. All loss terms are weighted equally; no adaptive weighting is used.

\emph{Attacker capability.} The attacker modifies $\lambda \to \lambda' = \lambda(1 + \delta)$ before training, where $\delta$ is the fractional perturbation. This models misconfiguration in automated pipelines, parameter drift in shared PINN libraries, or intentional tampering in collaborative settings.

\emph{Defender capability.} The defender observes only the trained model and its training loss $\mathcal{L}(\lambda')$. The defender does not have access to ground truth solutions or external validation data.

\emph{Silent failure.} We define a silent failure as $\mathcal{L}(\lambda') < \tau_{\text{loss}}$ and $\|\hat{u}_{\lambda'} - \hat{u}_{\lambda}\|_{L^2} / \|\hat{u}_{\lambda}\|_{L^2} > \tau_{\text{error}}$, where $\hat{u}_{\lambda}$ denotes the PINN solution trained with parameter $\lambda$, distinct from the analytical or numerical ground truth, $\tau_{\text{loss}} = 0.01$ (chosen as a round-number threshold near the clean cavity baseline loss of $\sim$0.003, representing a loss that would not raise concern in practice), and $\tau_{\text{error}} = 0.05$. This threshold is a heuristic anchored to the cavity baseline rather than normalized across PDEs, and cross-PDE loss scales differ substantially; we therefore treat threshold-based silent-failure counts as illustrative and rely primarily on the more robust observation that poisoned models can match or beat the clean-model loss while producing large solution differences (Section~5).

\emph{Detection difficulty ratio.} We define $R = (\|\hat{u}_{\lambda'} - \hat{u}_{\lambda}\|_{L^2} / \|\hat{u}_{\lambda}\|_{L^2}) / \mathcal{L}(\lambda')$. A high $R$ indicates that solution error is large relative to training loss. $R$ is the ratio of a dimensionless quantity to the training loss, whose absolute scale depends on the PDE; direct comparison across PDEs should be interpreted with care, as some variation reflects intrinsic differences in loss scale rather than vulnerability alone (Section~\ref{sec:discussion}).

\section{Experimental Setup}

\emph{PDE systems.} We test three systems of increasing complexity: (1)~Burgers' equation (1D, time-dependent, $\nu = 0.01/\pi$); (2)~Navier-Stokes lid-driven cavity (2D, steady, Re=100 and Re=400); (3)~Convection-diffusion (2D, time-dependent, $D=0.01$, Pe$\approx$33.5). For each PDE, the poisoned parameter ($\nu$, Re, or $D$) enters only the interior equation. Ground truths are finite-difference solutions (Burgers, conv-diff; an analytical Cole-Hopf solution exists for Burgers but we use FD for pipeline consistency) and the \citet{ghia1982high} benchmark (cavity). For the cavity, pressure is unconstrained (no gauge fixing) and no corner smoothing is applied, contributing to the clean baseline's 19.7\% $v$-velocity error against Ghia.

\emph{Architecture and training.} All PINNs use fully-connected networks with tanh activation and Glorot uniform initialization, implemented in DeepXDE \citep{lu2021deepxde} with PyTorch backend in float32. Training follows Adam ($\text{lr} = 10^{-3}$) with L-BFGS-B refinement (ftol $= 2.2 \times 10^{-15}$, gtol $= 10^{-5}$, maxiter $= 15{,}000$). All experiments ran on NVIDIA T4 and RTX 4000 Ada GPUs; total compute was approximately 300 GPU-hours.

\emph{Attack protocol.} For each PDE, we train a clean baseline with the true parameter, then train poisoned models with $\delta \in \{0.05, 0.10, 0.20, 0.50, 1.0\}$ (and $\delta = 2.0$ for Burgers and conv-diff). Bidirectional poisoning is tested for the cavity. Solution error is the relative $L^2$ norm between poisoned and clean predictions. Unless otherwise stated, solution error is measured relative to the clean PINN trained with the intended parameter, not directly against an analytical or numerical ground truth; the external comparison against the \citet{ghia1982high} benchmark in Section~5.6 is the one exception.

\section{Results}

\subsection{Parameter Poisoning Produces Silent Failures}

Table~\ref{tab:main} summarizes poisoning results across all PDE systems. Poisoned models consistently reach low training loss while producing incorrect solutions. The most severe case is the Navier-Stokes cavity at Re=400, where all five perturbation levels produce large solution errors ($0.26$--$0.71$ L2) with poisoned losses at or below the clean baseline. At $\delta = 1.0$ (Re doubled from 400 to 800), the poisoned model's training loss ($0.0071 \pm 0.0005$) is below the clean baseline ($0.0088 \pm 0.0003$), yet the solution is 71\% incorrect.

\begin{figure*}[t]
    \centering
    \includegraphics[width=\textwidth]{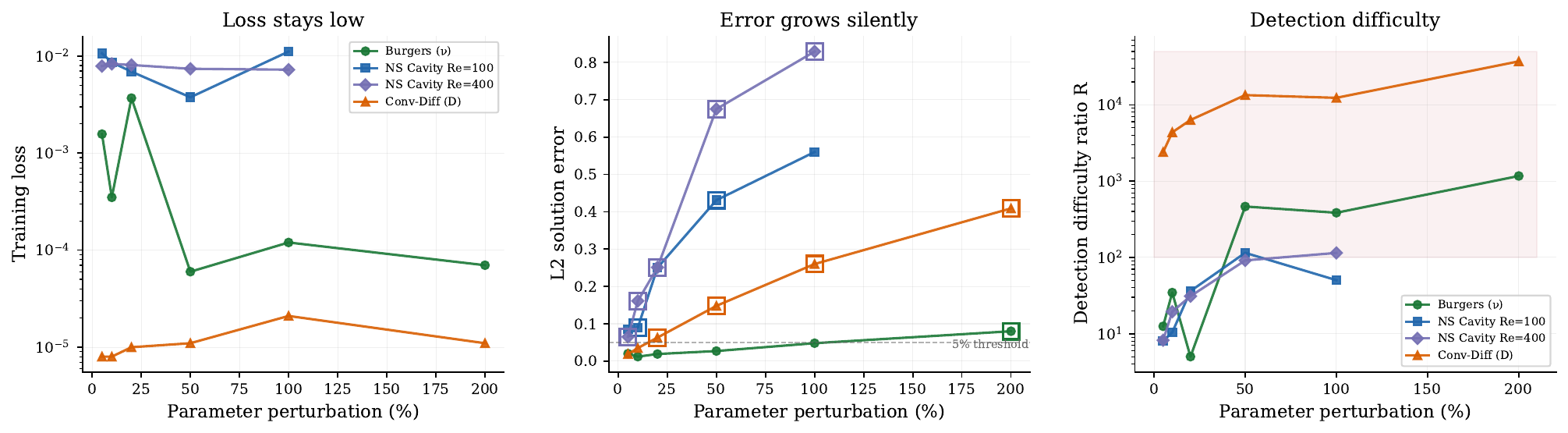}
    \caption{The Danger Zone: physics parameter poisoning across PDEs. \textbf{Left:} Training loss remains low regardless of perturbation magnitude. \textbf{Center:} Solution error grows silently; squares mark silent failures. \textbf{Right:} Detection difficulty ratio varies by orders of magnitude across PDE types.}
    \label{fig:dangerzone}
\end{figure*}

$R$ spans four orders of magnitude across PDE types (Figure~\ref{fig:dangerzone}). Burgers is the most resistant: error reaches only 8\% even at $\delta = +200\%$, and no perturbation in either direction produces a silent failure, consistent with spectral bias \citep{rahaman2019spectral} making smoother (higher-$\nu$) solutions easy to fit. The cavity is more vulnerable, especially at higher Re. Convection-diffusion is the worst case, with $R$ above $10^4$. Training losses in the $10^{-5}$ to $10^{-6}$ range approach the float32 precision floor; the large $R$ values for conv-diff are partly driven by small denominators and should be interpreted with caution. At Re=400, L2 error variance decreases with perturbation magnitude, suggesting that small perturbations land near bifurcation points where seed-dependent convergence dominates, while large perturbations overwhelm the solution structure consistently.

\begin{table*}[t]
\centering
\caption{Physics parameter poisoning results (mean $\pm$ std, $n=3$ seeds for cavity and Burgers; Conv-Diff single-seed except $\delta=200\%$). $R$ = detection difficulty ratio. Training losses are highly variable across seeds (in some cases $\sigma \geq \mu$), reflecting the stochasticity of PINN optimization rather than Gaussian sampling; L2 errors are more stable.}
\label{tab:main}
\small
\begin{tabular}{@{}llccc@{}}
\toprule
PDE & $\delta$ & Train Loss & L2 Error & $R$ \\
\midrule
Burgers ($\nu$) & Clean & 0.011$\pm$0.010 & --- & --- \\
& 5\% & 0.0006$\pm$0.0006 & 0.006$\pm$0.001 & 42$\pm$28 \\
& 10\% & 0.006$\pm$0.008 & 0.016$\pm$0.002 & 34$\pm$31 \\
& 20\% & 0.0009$\pm$0.0008 & 0.013$\pm$0.002 & 27$\pm$16 \\
& 50\% & 0.003$\pm$0.003 & 0.028$\pm$0.002 & 44$\pm$48 \\
& 100\% & 0.0007$\pm$0.0001 & 0.048$\pm$0.001 & 66$\pm$10 \\
& 200\% & 0.0001$\pm$0.0001 & 0.080$\pm$0.000 & 1413$\pm$1156 \\
\midrule
NS Cavity Re=100 & Clean & 0.0033$\pm$0.0006 & --- & --- \\
& 5\% & 0.0127$\pm$0.0066 & 0.241$\pm$0.090 & 28$\pm$23 \\
& 10\% & 0.0051$\pm$0.0013 & 0.131$\pm$0.070 & 24$\pm$11 \\
& 20\% & 0.0119$\pm$0.0055 & 0.248$\pm$0.139 & 19$\pm$4 \\
& 50\% & 0.0089$\pm$0.0035 & 0.279$\pm$0.014 & 36$\pm$12 \\
& 100\% & 0.0080$\pm$0.0025 & 0.415$\pm$0.112 & 53$\pm$8 \\
\midrule
NS Cavity Re=400 & Clean & 0.0088$\pm$0.0003 & --- & --- \\
& 5\% & 0.0089$\pm$0.0008 & 0.320$\pm$0.189 & 38$\pm$24 \\
& 10\% & 0.0082$\pm$0.0002 & 0.256$\pm$0.179 & 32$\pm$23 \\
& 20\% & 0.0084$\pm$0.0008 & 0.401$\pm$0.151 & 49$\pm$22 \\
& 50\% & 0.0089$\pm$0.0016 & 0.563$\pm$0.117 & 63$\pm$6 \\
& 100\% & 0.0071$\pm$0.0005 & 0.706$\pm$0.123 & 101$\pm$23 \\
\midrule
Conv-Diff ($D$) & Clean & 0.000006 & --- & --- \\
& 5\% & 0.000008 & 0.019 & 2{,}400 \\
& 10\% & 0.000008 & 0.035 & 4{,}362 \\
& 20\% & 0.000010 & 0.063 & 6{,}330 \\
& 50\% & 0.000011 & 0.148 & 13{,}427 \\
& 100\% & 0.000021 & 0.260 & 12{,}362 \\
& 200\% & 0.000025$\pm$0.000015 & 0.405$\pm$0.002 & 36{,}757$\pm$35{,}814 \\
\bottomrule
\end{tabular}
\end{table*}

\subsection{Adversarial Search}

To characterize worst-case vulnerability, we performed a coarse search over $\delta \in [-65\%, +100\%]$ for the cavity at Re=100, training a poisoned model at each $\delta$ and recording L2 error subject to the constraint $\mathcal{L} < \tau_{\text{loss}} = 0.01$. The worst case is $\delta = -60\%$ (Re=40): the model achieves loss 0.004 (well below $\tau$) while producing 128\% L2 error, substantially worse than the 71\% found in the fixed-$\delta$ sweep (Table~\ref{tab:main}). At this perturbation level, the flow regime is qualitatively different from Re=100 (L2 $>$ 1 indicates the prediction is worse than predicting zero), so the failure is severe rather than subtle. All negative perturbations produce feasible silent failures ($\mathcal{L} < \tau$ for every $\delta < 0$ tested), while positive perturbations above $+5\%$ frequently exceed $\tau$. This asymmetry reflects PINN spectral bias: lower Re produces smoother flow that the network fits easily with low loss, even though the solution differs dramatically from Re=100.

\subsection{Six Defenses: None Suffices}

We tested six candidate defenses.

\emph{PDE residual monitoring} produces a partial signal: the poisoned model satisfies its own (incorrect) PDE with high fidelity, but evaluating it against the \emph{stated} PDE (Re=100) produces a $3.3\times$ residual elevation (0.0112 vs 0.0034). The defender knows which equation was intended, so this check is available in principle. However, it requires a baseline expectation for the residual magnitude, and the signal diminishes at higher Re where training variance is large. The post-hoc loss sweep (Section~5.4) generalizes this idea: instead of checking the residual at a single stated parameter, it sweeps across parameter values and uses the location of the minimum as the diagnostic.

\emph{Same-parameter ensemble} (three independent poisoned models) fails: members disagree with each other more than they deviate from the clean baseline (ensemble std 0.470 vs error 0.175), reflecting training stochasticity rather than detection of poisoning. Disagreement is measured on velocity $(u, v)$ only; pressure is excluded due to gauge ambiguity.

\emph{Parameter-jitter ensemble} ($\pm2\%$, $\pm5\%$, $\pm10\%$ around stated Re): the poisoned model disagrees with the ensemble less than members disagree with each other (ratios 0.81$\times$, 0.44$\times$, 0.57$\times$). PINN training variance at nearby Reynolds numbers exceeds the systematic shift from poisoning, rendering this defense ineffective.

\emph{Inverse parameter recovery} (freeze network, learn $\lambda$) fails: recovered Re tracks initialization, not truth. The network absorbs data-fitting through its weights, leaving the parameter with negligible gradient pressure.

\emph{Incompressibility checking} ($\nabla \cdot \mathbf{u} = 0$) is Re-dependent: poisoned models show higher divergence at Re=100/150 but lower divergence at Re=400/800, making it unreliable as a standalone defense.

\emph{Loss elevation against clean baseline} is regime-dependent: detectable at Cavity Re=100 ($2.7\times$ elevation) but the poisoned loss falls below the clean baseline at Re=400, making it ineffective. Multi-seed architecture experiments show elevation ranges from $1.03\times$ (small network, essentially invisible) to $2.0\times$ (wide network, detectable). At moderate Re, the defense is further weakened by seed variance: roughly 38\% of seeds at Cavity Re=100 produce losses above $\tau_{\text{loss}} = 0.01$, blurring the boundary between detectable and silent.

\subsection{Post-Hoc Loss Landscape Defense}

The naive defenses above either operate within the corrupted physics framework or require a clean-baseline reference that may not be available. We propose a self-contained approach: \emph{post-hoc parameter-space loss sweep}. Given a trained model and a stated parameter value $\lambda_0$, evaluate the PDE residual loss at a range of parameter values $\lambda \in [\lambda_{\min}, \lambda_{\max}]$ without retraining. If the model was genuinely trained at $\lambda_0$, the residual loss should be minimized near $\lambda_0$. If the model was trained at a different $\lambda'$, the loss minimum will shift toward $\lambda'$. Crucially, this requires no clean-baseline comparison: the location of the loss minimum is informative on its own.

\begin{figure}[t]
    \centering
    \includegraphics[width=\columnwidth]{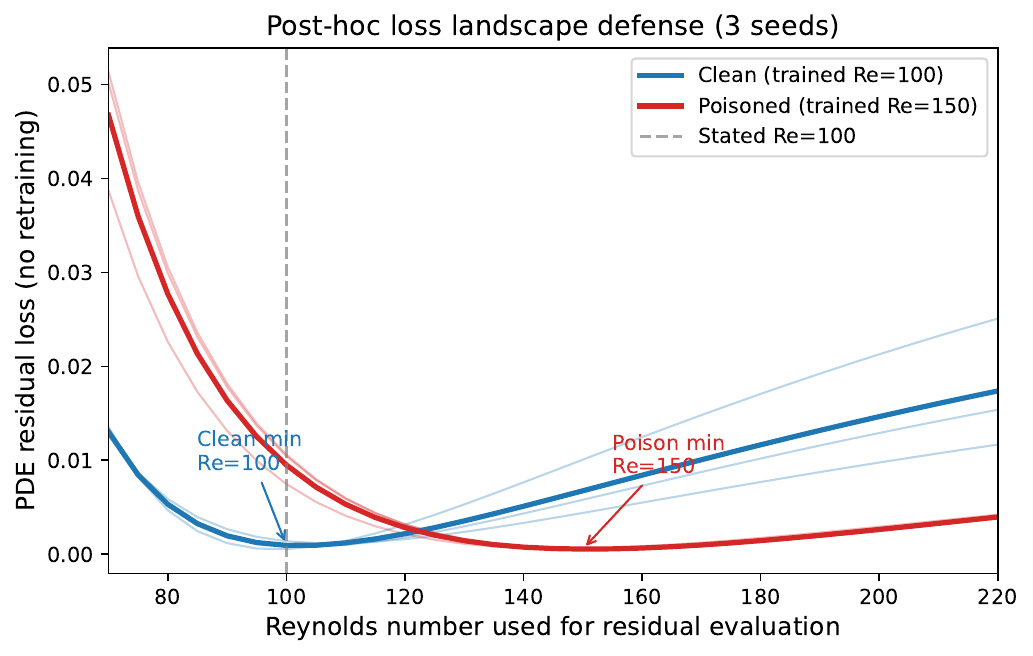}
    \caption{Post-hoc loss landscape defense ($n=3$ seeds). The PDE residual loss is evaluated at a range of Re values without retraining. Thin lines show individual seeds; thick lines show the mean. The clean model's mean minimum falls at Re=100 (the stated value); the poisoned model's mean minimum falls at Re=150 (the true training value). Individual seed minima range Re=100--105 (clean) and Re=145--155 (poisoned), with zero overlap.}
    \label{fig:sensitivity}
\end{figure}

Figure~\ref{fig:sensitivity} shows this defense across three seeds. The clean model's mean loss minimum falls at Re=100 (the stated value), while the poisoned model's mean minimum falls at Re=150 (the true training value). Individual seed minima range Re=100--105 (clean) and Re=145--155 (poisoned), with zero overlap. A defender checking whether the loss minimum falls within $\pm 25\%$ of the stated parameter would clear every clean model and flag every poisoned model.

The defense generalizes across all three PDE systems. For Burgers ($\delta = +100\%$, $\nu$ doubled), the sweep recovers the exact training viscosity. For convection-diffusion ($\delta = +200\%$, $D$ tripled), the sweep again recovers the exact training diffusivity. In both cases the clean model's loss minimum falls at the stated parameter value. The defense is not specific to the Navier-Stokes equations or to any particular parameter type.

Two supporting checks corroborate the finding. The gradient $\partial\mathcal{L}/\partial\text{Re}$ at stated Re=100 is $5.9\times$ steeper for the poisoned model, indicating it is far from its loss minimum. The spectral content of the velocity field differs: the poisoned model's high-to-low frequency energy ratio is $6.9\times$ that of the clean model. By contrast, mean kinetic energy is nearly identical (ratio 0.989), indicating bulk energy is not a useful discriminator.

This defense requires no retraining, no external reference data, and no modification to the training pipeline; only the ability to evaluate the PDE residual at alternative parameter values, which is trivially available in any PINN framework. The cost is $O(k)$ forward passes through the residual evaluator, where $k$ is the number of sweep points (we used $k=31$). The main limitation is that the defender must know which parameter might be corrupted and have a plausible range to sweep.

\subsection{Bidirectional Poisoning}

Poisoning works in both directions. We perturbed the cavity Reynolds number above and below Re=100. Six of eight non-trivial perturbations produce silent failures (Table~\ref{tab:bidir}). At Re=90, a 10\% reduction produces approximately 60\% solution error with a training loss (0.00324) comparable to the clean baseline (0.00334). The 3\% gap is within single-seed noise, but the training loss alone would not flag this model under any reasonable threshold. In a matched single-run comparison, negative Burgers perturbations produce slightly higher errors than positive perturbations, with higher training losses. No Burgers perturbation in either direction produces a silent failure, confirming resistance regardless of direction.

\begin{table}[t]
\centering
\caption{Bidirectional poisoning of the Navier-Stokes cavity (true Re=100). Single-seed. Six of the eight non-trivial perturbations are silent. The clearest case is Re=90, where a 10\% reduction yields 62.5\% L2 error at a training loss (0.00324) indistinguishable from the clean baseline (0.00334). The two non-silent cases, Re=105 and Re=200, carry the highest training losses in the sweep.}
\label{tab:bidir}
\small
\begin{tabular}{@{}cccc@{}}
\toprule
Re Used & Train Loss & L2 Error & Silent \\
\midrule
50 & 0.00392 & 0.134 & Yes \\
75 & 0.00389 & 0.274 & Yes \\
90 & 0.00324 & 0.625 & Yes \\
100 & 0.00334 & 0.000 & --- \\
105 & 0.01066 & 0.085 & No \\
110 & 0.00861 & 0.090 & Yes \\
120 & 0.00691 & 0.250 & Yes \\
150 & 0.00376 & 0.431 & Yes \\
200 & 0.01113 & 0.560 & No \\
\bottomrule
\end{tabular}
\end{table}

\subsection{External Validation}

We evaluated the poisoned cavity model (Re=150) against the \citet{ghia1982high} benchmark. The poisoned model shows approximately $2\times$ degradation in $u$-velocity accuracy and $1.7\times$ in $v$-velocity accuracy relative to the clean baseline. The clean baseline itself carries non-trivial error against Ghia (4.8\% in $u$, 19.7\% in $v$), reflecting known limitations of standard PINNs for cavity flow. The relative degradation from poisoning is clearly detectable: the external benchmark reveals a discrepancy that loss monitoring misses entirely.

\subsection{Architecture Independence and Multi-Seed Validation}

Five architectures (8.7K to 133K parameters, tanh and sinusoidal activations) all produce large solution errors under identical poisoning (Table~\ref{tab:arch_main}). Loss elevation varies from $1.03\times$ ($[64]\times3$ tanh) to $2.0\times$ ($[256]\times3$ tanh), meaning some architectures give the defender a detectable signal while others do not. The $[64]\times3$ network shows the largest L2 error ($1.30 \pm 0.41$) with negligible loss elevation; the poisoning is both severe and invisible to loss monitoring. The sinusoidal-activation network, which appeared uniquely silent when tested with one seed, shows $1.6\times$ elevation across three seeds. The single-seed result was not representative.

\begin{table*}[t]
\centering
\caption{Architecture independence of physics parameter poisoning (NS cavity, $\delta = 50\%$). Mean $\pm$ std, $n=3$ seeds.}
\label{tab:arch_main}
\small
\begin{tabular}{@{}lccccc@{}}
\toprule
Architecture & Params & Clean Loss & Poison Loss & L2 Error & $R$ \\
\midrule
$[128]\times5$, tanh & 66.8K & 0.0042$\pm$0.0010 & 0.0066$\pm$0.0013 & 0.405$\pm$0.057 & 66$\pm$25 \\
$[64]\times3$, tanh & 8.7K & 0.0147$\pm$0.0005 & 0.0151$\pm$0.0004 & 1.302$\pm$0.410 & 87$\pm$29 \\
$[256]\times3$, tanh & 133K & 0.0065$\pm$0.0007 & 0.0129$\pm$0.0004 & 0.265$\pm$0.017 & 21$\pm$1 \\
$[64]\times8$, tanh & 29.5K & 0.0064$\pm$0.0031 & 0.0076$\pm$0.0026 & 0.486$\pm$0.265 & 61$\pm$15 \\
$[128]\times5$, sin & 66.8K & 0.0047$\pm$0.0010 & 0.0075$\pm$0.0002 & 0.266$\pm$0.163 & 36$\pm$23 \\
\bottomrule
\end{tabular}
\end{table*}

Table~\ref{tab:main} reports multi-seed mean $\pm$ std ($n=3$) for all cavity and Burgers configurations. L2 errors are consistent across seeds for most cases (e.g., cavity Re=400 $\delta=100\%$: $0.706 \pm 0.123$). At Re=400, the multi-seed poisoned loss is below the clean baseline, confirming genuine silent failures. Training losses show higher variance at Re=100, where some seeds converge to qualitatively different solutions. At Re=400, L2 error variance decreases with perturbation magnitude: small perturbations land near bifurcation points where the outcome is seed-dependent, while large perturbations overwhelm the flow structure consistently.

\section{Discussion}
\label{sec:discussion}

The PINN loss measures only how well the model satisfies the equations it was given, so when those equations are incorrect a low loss certifies nothing more than that the model has learned the incorrect physics well. This appears to be the mechanism behind the silent failures we observe: the optimizer is faithful to a corrupted target, and the diagnostic the practitioner trusts is computed against that same corrupted target.

The detection difficulty ratio $R$ should be read within each PDE system rather than across them, since its four-order-of-magnitude spread is driven mostly by differences in absolute loss scale rather than by differences in how dangerous the poisoning is. Convection-diffusion illustrates the problem: its training losses sit near $10^{-5}$ to $10^{-6}$, so even a modest solution error divides by a tiny denominator and inflates $R$ into the thousands, making it appear the most vulnerable system when it is in fact the least. Normalizing by each system's own clean baseline removes this artifact. Defining $R' = (\text{relative } L^2 \text{ error})/(\mathcal{L}_{\text{poisoned}}/\mathcal{L}_{\text{clean}})$, the cross-PDE ordering inverts: at $\delta = 100\%$, $R'$ is $0.88$ for the cavity at Re=400, $0.17$ for the cavity at Re=100, and only $0.07$ for convection-diffusion, which may suggest that the cavity is the most vulnerable system and convection-diffusion the most benign. Burgers is the exception under either ratio; its training loss collapses well below its clean value at large perturbations, which makes $R'$ unstable ($0.75$ at $\delta = 100\%$) even though its absolute solution error never exceeds 8\%. We therefore report $R$ as a within-system severity indicator and treat $R'$ as the appropriate quantity for any cross-system claim, while noting that both ratios remain sensitive to the seed-to-seed variability of PINN training loss.

We frame these experiments as parameter poisoning to connect with the machine learning security literature, but the same results describe ordinary model misspecification, and the perturbation schedule is better understood as a sensitivity analysis than as an adversarial optimization. The practical concern is identical under either reading: a PINN user cannot distinguish correct from incorrect physics through training loss alone.

These failures matter because PINNs are being applied to gas turbine monitoring \citep{barimah2025scalable}, cardiac modeling \citep{buoso2021cardiac}, and turbomachinery flow reconstruction \citep{mcnichols2024pinn}, where parameter corruption, even accidental, could produce confident but incorrect predictions with no diagnostic warning. As pretrained PINN models begin to circulate through public repositories, the provenance of the encoded physics becomes a trust question in its own right.

The variation in detection difficulty across PDE types reflects both equation structure and PINN spectral bias. For Burgers, positive perturbations raise the viscosity and produce smoother solutions that the network fits easily while staying close to the clean solution in $L^2$ norm. For convection-diffusion, the linearity of the operator means parameter changes produce proportional solution changes with little loss penalty. For the cavity, the nonlinear coupling between velocity components produces complex flow structures that change substantially with Reynolds number, which may explain why it appears most vulnerable once the loss-scale artifact is removed. The bidirectional asymmetry at Re=100, where a 10\% decrease in Re produces a qualitatively larger error than a 10\% increase, likely reflects the same spectral bias: the network converges readily to the smoother Re=90 solution and reproduces it at low loss even though it differs substantially from the Re=100 reference.

\section{Limitations}

Table~\ref{tab:main} reports multi-seed ($n=3$) for cavity and Burgers; other tables remain single-seed. Ideally $n \geq 5$ seeds would be used. The six defenses test increasingly sophisticated but still naive formulations; Bayesian PINNs \citep{yang2021bpinn} and adaptive loss weighting were not tested. All architectures are standard feedforward networks; Fourier neural operators and PirateNet are untested. We sweep $\delta$ over fixed values in Table~\ref{tab:main} and additionally perform a coarse adversarial search over $\delta \in [-65\%, +100\%]$; the worst case found ($\delta = -60\%$, Re=40, L2=1.28) may not be the global optimum. The post-hoc loss landscape defense has been validated across all three PDE systems and across three random seeds for the cavity case, with zero false positives or false negatives.

\section{Conclusion}

Corrupted PDE coefficients produce PINNs that converge to low training loss while giving incorrect solutions. This failure mode, whether interpreted as parameter poisoning or model misspecification, is systematic, bidirectional, PDE-dependent, and architecture-independent. Six naive defenses provide only partial signals; none works across all regimes. Post-hoc loss landscape analysis, sweeping the PDE residual across parameter values without retraining, successfully reveals the true training parameter across all three PDE systems tested. This defense requires no external data, no retraining, and no modification to the training pipeline. Low training loss does not certify physical correctness.

\bibliographystyle{plainnat}
\bibliography{references}

\end{document}